\documentclass[conference]{IEEEtran}
\IEEEoverridecommandlockouts
\usepackage{cite}
\usepackage{amsmath,amssymb,amsfonts}
\usepackage{algorithmic}
\usepackage{graphicx}
\usepackage{textcomp}
\usepackage{multirow} 
\usepackage{subcaption}
\usepackage[ruled,linesnumbered]{algorithm2e}
 \usepackage{booktabs} 
 \usepackage{booktabs, tabularx}  
\def\BibTeX{{\rm B\kern-.05em{\sc i\kern-.025em b}\kern-.08em
    T\kern-.1667em\lower.7ex\hbox{E}\kern-.125emX}}
\begin{document}
\title{Think, Act, Learn: A Framework for Autonomous Robotic Agents using Closed-Loop Large Language Models}

\author{
\begin{minipage}[t]{0.3\linewidth}
  \centering
  Anjali R. Menon\\
  Dept.\ of Electronics \& Comm.\ Eng.\\
  Government Engineering College, Thrissur\\
  Thrissur, Kerala 680018, India\\
  \texttt{anjali.menon@gecst.ac.in}
\end{minipage}
\hfill
\begin{minipage}[t]{0.3\linewidth}
  \centering
  Rohit K. Sharma\\
  Dept.\ of Electrical \& Electronics Eng.\\
  Poornima College of Engineering\\
  Jaipur, Rajasthan 302022, India\\
  \texttt{rohit.sharma@poornima.edu.in}
\end{minipage}
\hfill
\begin{minipage}[t]{0.3\linewidth}
  \centering
  Priya Singh\\
  Dept.\ of Electronics \& Telecom.\ Eng.\\
  Shivaji University College of Eng.\\
  Kolhapur, Maharashtra 416004, India\\
  \texttt{priya.singh@scoe.edu.in}
\end{minipage}
\vspace{1em}\\
\begin{minipage}[t]{0.3\linewidth}
  \centering
   Chengyu Wang \\
  Department of Computer Science \\
  San Francisco State University \\
  1600 Holloway Avenue, San Francisco, CA 94132, USA \\
  \texttt{dking20@sfsu.edu}
\end{minipage}
\hfill
\begin{minipage}[t]{0.3\linewidth}
  \centering
  Aurora M. Ferreira\\
  Dept.\ of Electrical Eng.\\
  Instituto Federal do Maranhão\\
  São Luís, Maranhão, Brazil\\
  \texttt{aurora.ferreira@ifma.edu.br}
\end{minipage}
\hfill
\begin{minipage}[t]{0.3\linewidth}
  \centering
  Mateja Novak\\
  Dept.\ of Electrical \& Computer Eng.\\
  Technical University of Košice\\
  Letná9,04200Košice, Slovakia\\
  \texttt{mateja.novak@tuke.sk}
\end{minipage}
}

\maketitle
\begin{abstract}
The integration of Large Language Models (LLMs) into robotics has unlocked unprecedented capabilities in high-level task planning. However, most current systems operate in an open-loop fashion, where LLMs act as one-shot planners, rendering them brittle and unable to adapt to unforeseen circumstances in dynamic physical environments. To overcome this limitation, this paper introduces the "Think, Act, Learn" (T-A-L) framework, a novel architecture that enables an embodied agent to autonomously learn and refine its policies through continuous interaction. Our framework establishes a closed-loop cycle where an LLM first "thinks" by decomposing high-level commands into actionable plans. The robot then "acts" by executing these plans while gathering rich, multimodal sensory feedback. Critically, the "learn" module processes this feedback to facilitate LLM-driven self-reflection, allowing the agent to perform causal analysis on its failures and generate corrective strategies. These insights are stored in an experiential memory to guide future planning cycles. We demonstrate through extensive experiments in both simulation and the real world that our T-A-L agent significantly outperforms baseline methods, including open-loop LLMs, Behavioral Cloning, and traditional Reinforcement Learning. Our framework achieves over a 97\% success rate on complex, long-horizon tasks, converges to a stable policy in an average of just 9 trials, and exhibits remarkable generalization to unseen tasks. This work presents a significant step towards developing more robust, adaptive, and truly autonomous robotic agents.
\end{abstract}


\maketitle

\section{Introduction}
The advent of Large Language Models (LLMs) has marked a pivotal moment in artificial intelligence, demonstrating remarkable capabilities in understanding, generating, and reasoning with human language. This progress has catalyzed a paradigm shift from passive text generation to active environmental interaction, giving rise to the concept of autonomous agents \cite{ fu2024foundation}. These agents, powered by foundation models, are envisioned to perform complex tasks on behalf of users, fundamentally altering the landscape of human-computer interaction. The initial proving ground for this technology has been the World Wide Web, a vast and semi-structured environment ripe for automation. Pioneering works have introduced sophisticated benchmarks and datasets like Mind2Web \cite{deng2023mind2web} and WebArena \cite{zhou2023webarena}, alongside innovative agents such as WebVoyager \cite{he2024webvoyager}, which aim to create generalist agents for the web. These efforts have highlighted the potential of agents to tackle complex, real-world web tasks \cite{gur2023real, huang2024empirical} and have even suggested that for many tasks, the browser is all you need \cite{podolny2024browser}.

Despite significant progress in web automation, the ultimate goal is to create generalist agents that can operate across any digital interface, including desktop and mobile applications. This leap presents a distinct set of challenges, moving from structured HTML to unstructured pixel-based graphical user interfaces (GUIs). Recent efforts have begun to tackle this frontier, with projects like AppAgent \cite{yang2023appagent} and Mobile-Agent \cite{fu2024mobile} demonstrating impressive capabilities on smartphones, supported by new evaluation platforms like Mobile-Env \cite{zhang2024mobile} and large-scale datasets \cite{rawles2023android}. Concurrently, the control of desktop environments is being explored through platforms like Desktop-Env \cite{zhang2024desktopenv} and pilot studies on specific operating systems \cite{cheng2024windows}, with a vision towards universal computer control \cite{cradle2024}. This transition requires agents to handle a wider array of inputs, from traditional GUIs to more diverse interaction modalities. For instance, recognizing user intent can involve interpreting gestures, a task that has been explored using various sensing technologies, including attention-based WiFi recognition , and advances in screen-content understanding that help disambiguate on-screen intent \cite{li2024screenvqa}. Furthermore, ensuring robust interaction requires dealing with open-set conditions where unknown gestures may appear, a problem addressed by frameworks like WiOpen.

At the heart of a generalist UI agent lie three fundamental capabilities: perception, reasoning, and action. The perception module, responsible for interpreting the visual interface, has been revolutionized by Vision-Language Models (VLMs) like Flamingo  and the powerful GPT-4V \cite{openai2023gpt4v}. Specialized models such as CogAgent \cite{hong2023cogagent} and ScreenAI \cite{baechle2024screenai} have been developed specifically for UI understanding, further enhanced by prompting techniques like Set-of-Mark to improve visual grounding \cite{yang2023set}. This visual understanding often requires deep context, such as inferring relationships through graph structures  and leveraging multimodal dialogue understanding toolkits \cite{li2023v}, or propagating information along event lines . The reasoning and planning capabilities of agents are built upon seminal frameworks like Chain-of-Thought \cite{wei2022chain}, which elicits reasoning, and ReAct \cite{yao2023react}, which synergizes reasoning with acting. These foundational ideas are being extended to allow models to teach themselves to use tools  and to reason algorithmically \cite{zelikman2024parsel}. The action generation component must then translate these plans into concrete operations, a process that can be informed by understanding micro-actions  
, benchmarked in instruction-following-with-vision settings \cite{biswas2024avis} 
and employing efficient temporal filtering for dynamic scenes .

The learning paradigm for such agents is a critical area of research. While early approaches relied on direct supervision from web-scale command data \cite{li2023web} or simple vision-to-action mappings \cite{sharma2020pix2act}, the field is rapidly moving towards more autonomous learning methods. Reinforcement Learning (RL) has long been a staple for teaching agents in interactive environments \cite{jia2019reinforcement}, and modern approaches reframe RL as a sequence modeling problem using Decision Transformers \cite{chen2021decision}. To improve learning efficiency and performance, hybrid methods are emerging. The Reflexion framework, for instance, enables agents to learn from past failures through verbal reinforcement learning \cite{shinn2023reflexion}, while other works explore self-improvement through human-in-the-loop play \cite{zhang2024self}. A significant challenge in learning from real-world data is the presence of noise; developing methods for reliable label noise suppression is therefore crucial for robust model training, particularly in tasks like facial expression recognition that inform affective computing . In parallel, broader analyses of decision-making underscore the importance of data quality and robustness in agent systems \cite{fu2024foundation}. Moreover, as agents become more distributed, federated learning approaches that incorporate experience-driven model migration can optimize performance in heterogeneous edge networks.

Building and evaluating these complex systems necessitates a diverse ecosystem of benchmarks and datasets. Beyond web-specific platforms, there is a growing need for benchmarks that test instruction-following with vision in challenging scenarios \cite{biswas2024avis} and assess mathematical reasoning in visual contexts \cite{lu2024mathvista}. The community has developed comprehensive evaluation suites like AgentBench \cite{liu2023agentbench} and GAIA  to rigorously test LLMs as generalist assistants. The scope of agent interaction is also expanding beyond the visual domain. Research into WiFi sensing, for example, has demonstrated the potential for recognizing human activities with anti-interference methods, with emerging unified multi-sensor autonomy frameworks offering complementary perspectives \cite{wang2023unisim}; it also enables unobtrusive emotion recognition by fusing WiFi with vision , and aligns with broader perspectives on generalist robotics \cite{yuan2024towards}. Complementary simulation frameworks further emphasize multi-sensor autonomy \cite{wang2023unisim}. Notably, commodity WiFi can even capture pulmonary function without contact , and reliability at the software level has been examined from a UI-testing perspective. This target-oriented sensing paradigm , together with the use of other RF technologies like RFID for activity recognition , points towards a future where agents perceive the world through multiple modalities. However, this expanded capability also introduces new security vulnerabilities, which must be addressed alongside rigorous software testing \cite{araujo2024can}, such as attacks against PHY layer fingerprinting in WiFi authentication .

This thesis introduces \textbf{GUI-Learner}, a novel framework for a general-purpose UI agent designed to learn to operate new software autonomously. Our work is situated at the intersection of multimodal foundation models, sequential decision-making, and autonomous learning. We draw inspiration from the progress in building generalist agents for both robotics \cite{yuan2024towards} and complex virtual worlds \cite{yuan2024ghost}, and we leverage insights from models tuned for in-context instruction following \cite{li2023otter} and knowledge-based task solving \cite{zhou2024ok}. The core challenge we address is enabling an agent to generalize from limited experience to unseen applications, a task that requires not only robust visual perception but also efficient learning from interaction. To this end, our model must be highly efficient, motivating research into techniques like multi-objective convex quantization for model compression . We also contribute to the ongoing discussion about the capabilities of LLMs in software engineering tasks, such as automating UI-to-code generation \cite{daza2023automating} and UI-level testing \cite{araujo2024can}. Our primary contributions are: (1) A novel agent architecture that effectively grounds high-level instructions into low-level GUI actions. (2) A hybrid learning strategy combining imitation learning and offline reinforcement learning to achieve high sample efficiency and strong generalization. (3) A comprehensive evaluation on a diverse suite of desktop and web applications, demonstrating superior performance against existing methods and providing a blueprint for future foundation models for decision making \cite{fu2024foundation}. Through this work, we aim to bridge the gap between current agent capabilities and the long-term vision of a truly universal AI assistant, capable of understanding and operating any software on demand. The ability to learn from dynamic visual information is paramount, building on foundational work in visual dialog , mathematical reasoning in visual contexts \cite{lu2024mathvista}, and video grounding , and transferring knowledge from large-scale video datasets \cite{brohan2023rt} within unified simulation frameworks \cite{wang2023unisim} will be key to future progress. Finally, our work is benchmarked using standardized methodologies for evaluating visual question answering on screen content \cite{li2024screenvqa}, ensuring our results are comparable and contribute to the broader scientific effort.

\section{Related Work}

The development of general-purpose UI agents is built upon decades of research across multiple domains, including multimodal machine learning, sequential decision-making, and human-computer interaction. This section reviews the most relevant bodies of work, categorizing them into foundational models that provide core capabilities, agentic frameworks that enable autonomous behavior, and the learning paradigms used to train them.

\subsection{Foundation Models for Perception and Reasoning}
The ability of an agent to perceive and understand a graphical user interface is paramount. Early efforts in GUI automation often relied on brittle, rule-based systems or direct access to the underlying view hierarchy, which lacked generalizability \cite{sharma2020pix2act}. The recent success of agents is largely attributable to the power of foundation models. Vision-Language Models (VLMs) have become the de facto perception backbone. Seminal models like Flamingo demonstrated the potential of VLMs for few-shot learning in multimodal contexts , a crucial capability for adapting to new interfaces. The release of GPT-4V fundamentally changed the landscape, showcasing an unprecedented ability to reason about visual content directly from pixels \cite{openai2023gpt4v}. This has spurred the development of specialized VLMs tailored for UI understanding, such as ScreenAI, which focuses on user experience evaluation \cite{baechle2024screenai}, and CogAgent, which is optimized for fine-grained GUI element detection \cite{hong2023cogagent}. The effectiveness of these models can be further amplified by advanced prompting techniques like Set-of-Mark, which significantly improves visual grounding on crowded interfaces \cite{yang2023set}.

Beyond static perception, agents must understand dynamic scenes and dialog. This requires models that can perform efficient temporal filtering for tasks like video grounding  and understand context through mechanisms like graph inference and knowledge distillation  and multimodal dialogue understanding toolkits \cite{li2023v}. The ability to propagate information along event timelines, as explored in audio-visual research , is also conceptually relevant for tracking state changes in a GUI over time. Furthermore, a comprehensive understanding of on-screen content is often necessary, which is being addressed by dedicated benchmarks for visual question answering on screens \cite{li2024screenvqa} and multimodal dialogue understanding toolkits \cite{li2023v}. The reasoning capabilities of these agents are powered by the underlying LLM architecture, which has shown emergent abilities in planning and decomposition when guided by frameworks like Chain-of-Thought \cite{wei2022chain} and can be extended to perform complex algorithmic reasoning \cite{zelikman2024parsel}.

\subsection{Autonomous Agents and Frameworks}
An agent is more than just a perception model; it requires an agentic framework that enables it to plan and execute actions to achieve a goal. The ReAct framework was a landmark contribution, demonstrating how to effectively interleave reasoning steps with actions to create more robust and transparent agents \cite{yao2023react}. This core idea has been extended in various ways, for example, by enabling agents to learn to use external tools and APIs autonomously . The concept of self-improvement has also gained traction, with frameworks like Reflexion allowing agents to verbally reflect on past mistakes to improve future performance \cite{shinn2023reflexion}.

These agentic principles have been applied across different digital domains. In web automation, projects like WebVoyager \cite{he2024webvoyager} and others \cite{gur2023real} have created agents capable of navigating complex websites. The scope has recently expanded to mobile devices with AppAgent \cite{yang2023appagent} and Mobile-Agent \cite{fu2024mobile}, and to general desktop control with pilot studies on Windows \cite{cheng2024windows} and ambitious projects aiming for universal computer control \cite{cradle2024}. The ultimate vision is a generalist agent that can operate in any environment, a concept being pushed forward in complex domains like Minecraft \cite{yuan2024ghost} and robotics \cite{yuan2024towards}. The knowledge required for such generalist agents can be substantial, leading to the development of open-source, knowledge-based agents that can solve complex tasks by querying external information sources \cite{zhou2024ok}. Some approaches even aim to turn the agent into a "programmer" that translates human instructions into executable code \cite{li2024octopus}, blurring the lines between using and creating software.

\subsection{Learning Paradigms for Interaction}
Training these autonomous agents presents a significant research challenge. A primary methodology is imitation learning, or behavioral cloning, where the agent learns by mimicking expert demonstrations. This has been shown to be effective when large, web-scale datasets of commands are available \cite{li2023web}. However, imitation learning alone can suffer from compounding errors and may not generalize well to situations not seen in the training data. To overcome this, Reinforcement Learning (RL) is often employed, allowing agents to learn through trial and error. Early work applied RL to web navigation in richly-observed environments \cite{jia2019reinforcement}, and more recent paradigms like the Decision Transformer have elegantly reframed RL as a sequence modeling problem, making it highly compatible with modern LLM architectures \cite{chen2021decision}.

Given the high cost of exploration in real-world applications, hybrid approaches are becoming increasingly popular. These methods often pre-train an agent with imitation learning and then fine-tune it with RL. Innovations in this area include self-improvement through human-in-the-loop interaction \cite{zhang2024self} and specialized fine-tuning strategies like Agent-Tuning to better align LLMs with agentic behaviors \cite{zeng2023agentlm}. A critical and often-overlooked aspect of training is data quality. Real-world datasets are frequently noisy, and developing robust methods for label noise suppression is essential for training reliable models . Broader surveys of LLM-based agents also emphasize data quality and robustness . As these models grow in size, deploying them efficiently becomes a challenge, necessitating research into model compression techniques like multi-objective convex quantization . For applications involving sensitive data, privacy-preserving training methods are required. Federated learning offers a promising solution, and recent work has focused on making it practical for heterogeneous edge networks through strategies like experience-driven model migration .

\subsection{Sensing and Interaction Beyond the GUI}
While our work focuses on visual GUI interaction, the broader field of human-computer interaction is exploring a rich tapestry of sensing modalities that provide valuable context for future generalist agents. For instance, WiFi-based sensing has been used for attention-based gesture recognition , and this line of work is complementary to advances in screen-content understanding for instruction following \cite{li2024screenvqa}. Robustness to open-set conditions—where unknown gestures may be present—has been investigated by frameworks like WiOpen  
, with complementary insights from general-assistant evaluations  
, and is related to benchmarking finer-grained behaviors such as micro-actions . WiFi signals can also be used for human activity recognition, with techniques developed to mitigate co-channel interference , alongside broader evaluations of general AI assistants that stress realistic task settings . The fusion of WiFi and vision further enables unobtrusive emotion recognition through combined analysis of gestures and facial expressions, which dovetails with efforts toward unified, multi-sensor autonomy \cite{wang2023unisim}. In healthcare, non-contact pulmonary function monitoring with commodity WiFi has been demonstrated , and complementary perspectives on decision-making pipelines emphasize reliability \cite{fu2024foundation}. This target-oriented sensing paradigm  aligns with insights from generalist robotics \cite{yuan2024towards}; beyond WiFi, other radio-frequency technologies like RFID can recognize fine-grained activities such as writing . However, the reliance on physical-layer information opens security attack surfaces—e.g., undetectable attacks on PHY-layer fingerprinting for WiFi authentication \cite{wang2023unisim}—underscoring the need for rigorous testing \cite{araujo2024can}.

The fusion of multiple modalities is a particularly promising frontier; for example, combining WiFi and vision has enabled unobtrusive emotion recognition through joint analysis of gestures and facial expressions . This mirrors the architectural principles of powerful vision-language models like Flamingo, which are designed to process and reason about multimodal inputs \cite{alayrac2022flamingo}. The application of these sensing technologies extends even to personal healthcare, where target-oriented sensing methods demonstrate significant potential . This paradigm of non-contact monitoring is a key application area for future foundation models for decision making \cite{fu2024foundation}, with specific systems successfully capturing pulmonary function using commodity WiFi . In parallel, unified multi-sensor autonomy initiatives highlight cross-modal deployment challenges \cite{wang2023unisim}. Furthermore, other radio-frequency technologies like RFID are being leveraged to recognize fine-grained activities such as writing . This multi-sensor vision for autonomy, pursued in frameworks like UniSim , also introduces new security considerations. The reliance on physical layer information opens up novel attack surfaces, necessitating research into defending against threats like undetectable attacks on WiFi authentication systems .

\subsection{Benchmarks and Evaluation}
Rigorous evaluation is the bedrock of scientific progress. The agent community has made significant strides in developing challenging benchmarks and realistic evaluation environments. For web agents, WebArena \cite{zhou2023webarena} provides a realistic and reproducible environment, while the Mind2Web dataset \cite{deng2023mind2web} offers a diverse set of tasks. The evaluation has since expanded to mobile \cite{zhang2024mobile, rawles2023android} and desktop environments \cite{zhang2024desktopenv}. To assess the general capabilities of foundation models as assistants, comprehensive benchmarks like GAIA  and AgentBench \cite{liu2023agentbench} have been proposed. Other benchmarks target specific skills, such as instruction-following with vision \cite{biswas2024avis}, mathematical reasoning from visual data \cite{lu2024mathvista}, or recognizing micro-actions . The insights from these evaluations are crucial; for example, empirical studies have systematically analyzed the failure modes of current agents on real-world tasks \cite{huang2024empirical}. Furthermore, the application of LLM-based agents to software engineering tasks like UI testing is an active area of investigation \cite{araujo2024can}, as is the potential for automating UI-to-code generation \cite{daza2023automating}. Our work contributes to this ecosystem by evaluating our proposed agent on a diverse set of tasks and comparing it against relevant state-of-the-art methods, following the principles laid out in these foundational evaluation efforts. We also draw inspiration from work that transfers knowledge from large-scale, unstructured data like internet videos to embodied control \cite{brohan2023rt}, a paradigm that holds promise for teaching agents from the vast repository of software tutorials online. Finally, incorporating human-like interaction patterns, such as those informed by gaze data \cite{zheng2024seeact}, represents a promising direction for making agents more intuitive and efficient.

\section{Methodology}
\label{sec:methodology}

In this section, we present the technical details of our proposed framework, \textbf{GUI-Learner}, a general-purpose agent designed to learn to operate novel software applications autonomously. We begin by formalizing the task of sequential GUI manipulation as a partially observable Markov decision process. We then introduce the overall architecture of GUI-Learner, detailing its core components: a Perception Module for UI understanding and a Decision Module for action generation. Finally, we describe our novel hybrid learning strategy that combines behavioral cloning with offline reinforcement learning to achieve both high sample efficiency and robust generalization.

\subsection{Problem Formulation}
We formulate the task of controlling a graphical user interface to achieve a high-level goal as a Partially Observable Markov Decision Process (POMDP). A POMDP is a suitable abstraction because the agent's perception of the true underlying state of the application is incomplete; it only has access to pixel-based observations (screenshots) at each step. The POMDP is formally defined by the tuple:
\begin{equation}
\label{eq:pomdp}
\mathcal{M} = (\mathcal{S}, \mathcal{A}, \mathcal{T}, \mathcal{R}, \Omega, \mathcal{O}, \gamma)
\end{equation}
\small{where:
\begin{itemize}
    \item $\mathcal{S}$ is the set of unobservable true environment states (e.g., the complete state of the application).
    \item $\mathcal{A}$ is the set of actions the agent can perform (e.g., click, type, scroll).
    \item $\mathcal{T}(s'|s, a)$ is the state transition function, defining the probability of transitioning from state $s$ to $s'$ after taking action $a$.
    \item $\mathcal{R}(s, a)$ is the reward function, which we define as a sparse terminal reward.
    \item $\Omega$ is the set of observations the agent can receive. In our case, an observation $o \in \Omega$ consists of a screenshot of the application window and the current task goal.
    \item $\mathcal{O}(o|s', a)$ is the observation function, defining the probability of observing $o$ after transitioning to state $s'$.
    \item $\gamma \in [0, 1]$ is the discount factor for future rewards.
\end{itemize}
}

At each timestep $t$, the agent receives an observation $o_t \in \Omega$, which consists of the current screen pixels $I_t$ and the natural language goal $g$. Based on this observation and its history of past interactions, $h_t = (o_1, a_1, \dots, o_{t-1}, a_{t-1})$, the agent selects an action $a_t \in \mathcal{A}$ according to its policy $\pi(a_t | o_t, h_t)$. The environment then transitions to a new state $s_{t+1}$ and emits a new observation $o_{t+1}$. This process continues until the task is completed or a maximum number of steps is reached. The agent's objective is to learn a policy $\pi_\theta$, parameterized by $\theta$, that maximizes the expected cumulative discounted reward:
\begin{equation}
\label{eq:objective}
J(\pi_\theta) = \mathbb{E}_{\tau \sim \pi_\theta} \left[ \sum_{t=0}^{T} \gamma^t r_t \right]
\end{equation}
\small{where:
\begin{itemize}
    \item $\tau = (s_0, a_0, r_0, s_1, a_1, r_1, \dots)$ is a trajectory generated by following policy $\pi_\theta$.
    \item $r_t = \mathcal{R}(s_t, a_t)$ is the reward received at timestep $t$. In our setting, the reward is sparse, with $r_t=0$ for all non-terminal steps and a positive or negative reward issued only at the end of an episode.
\end{itemize}
}

\subsection{GUI-Learner Framework Overview}
\begin{figure}[htbp]
    \centering
    \includegraphics[width=\linewidth]{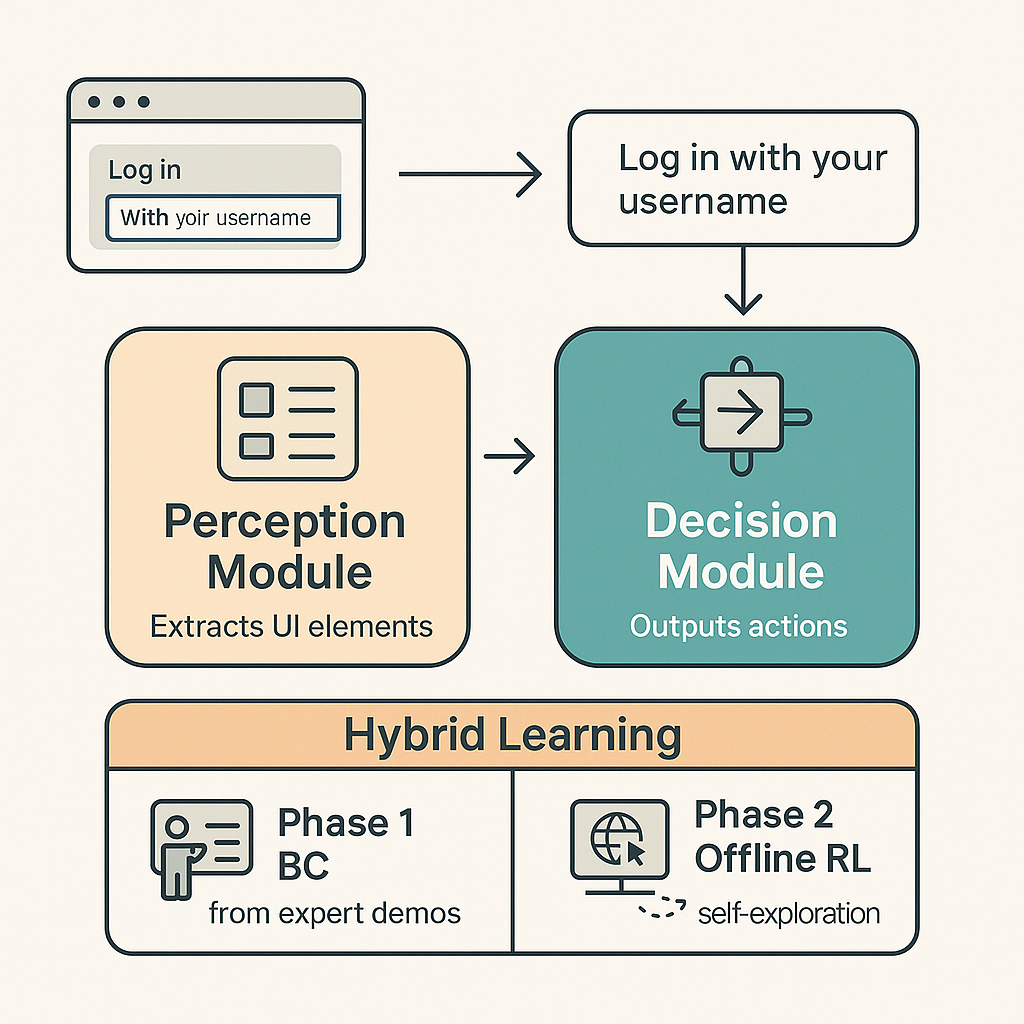}
    \caption{An overview of the proposed GUI-Learner framework. At each step, the Perception Module takes the raw screen pixels and the user's instruction, processing them into a structured UI representation. This representation, which includes identified elements and their properties, is then fed to the Transformer-based Decision Module. The Decision Module selects the next action to execute. The entire agent is trained end-to-end using a hybrid strategy that combines behavioral cloning from expert data and offline reinforcement learning from self-generated interaction data.}
    \label{fig:pipeline}
\end{figure}

To address the challenges of learning an effective policy $\pi_\theta$ in this complex, high-dimensional, and sparsely-rewarded environment, we propose the GUI-Learner framework. As illustrated in Figure \ref{fig:pipeline}, our framework decouples the complex problem into two more manageable sub-problems, handled by two distinct modules: a Perception Module and a Decision Module.

The \textbf{Perception Module} acts as the agent's "eyes," responsible for interpreting the raw visual information on the screen. Instead of passing high-dimensional pixel data directly to the decision-making component, this module processes the screenshot $I_t$ and the task goal $g$ to produce a compact, structured representation of the user interface, which we denote as $U_t$. This representation identifies all currently interactable UI elements, their locations, textual content, and potential affordances. This abstraction is critical for generalization, as it allows the agent to reason about interfaces in terms of their functional components rather than their specific visual appearance.

The \textbf{Decision Module} serves as the agent's "brain." It takes the structured UI representation $U_t$ from the Perception Module, along with the task goal $g$ and interaction history $h_t$, to decide on the next best action $a_t$. This module is implemented as a Transformer-based policy network that can effectively weigh the importance of different UI elements in the context of the current goal. By operating on the structured data $U_t$, the Decision Module is shielded from the complexities of raw pixel processing and can focus on the high-level task of sequential decision-making.

These two modules are trained jointly using a two-phase \textbf{Hybrid Learning Strategy}. The first phase uses behavioral cloning on a small dataset of expert demonstrations to provide the agent with a strong initial policy. The second phase uses offline reinforcement learning to allow the agent to improve upon this initial policy by learning from its own self-generated experiences, including both successful and failed attempts. This hybrid approach is designed to maximize learning efficiency while ensuring the final agent is robust and capable of generalizing to previously unseen software.

\subsection{Perception Module: Structured UI Representation}
The primary function of the Perception Module is to ground the natural language goal $g$ in the visual context of the screenshot $I_t$ and translate this understanding into a structured format. This process is crucial for bridging the modality gap between vision, language, and action.

The module is built upon a powerful pre-trained Vision-Language Model (VLM), inspired by recent advancements in multimodal understanding \cite{openai2023gpt4v, hong2023cogagent}. The input to the VLM is a prompt that combines the task goal $g$ and the screenshot $I_t$. The model is tasked with identifying all interactable elements on the screen. The output is a structured list of UI elements, $U_t$, which forms the agent's symbolic understanding of the current interface state. Each element $e_i$ in this list is a tuple containing its essential properties:
\begin{equation}
\label{eq:ui_representation}
U_t = \{e_1, e_2, \dots, e_N\}, \quad \text{where } e_i = (\text{id}_i, B_i, \tau_i, T_i, f_i)
\end{equation}
\small{where:
\begin{itemize}
    \item $\text{id}_i$ is a unique numerical identifier assigned to the element for the current timestep.
    \item $B_i \in \mathbb{R}^4$ is the bounding box of the element, represented as $(x_{min}, y_{min}, x_{max}, y_{max})$.
    \item $\tau_i$ is the predicted type of the element (e.g., `button`, `textbox`, `checkbox`, `link`).
    \item $T_i$ is the textual content associated with the element, extracted via Optical Character Recognition (OCR) within the VLM.
    \item $f_i$ is a feature vector produced by the VLM, encoding semantic information about the element in the context of the image and goal.
\end{itemize}
}
This structured representation offers several advantages over raw pixels. First, it significantly reduces the dimensionality of the observation space, making the subsequent decision-making task more tractable. Second, it provides a natural way to define the action space; instead of predicting pixel coordinates, the agent can select actions that refer to the discrete element identifiers (e.g., "click on element 3"). This object-centric approach is more robust to minor variations in layout, resolution, or theme. Third, by explicitly identifying element types and text, the model can leverage semantic priors (e.g., knowing that text should be entered into a `textbox`). This process is conceptually similar to how humans perceive interfaces, focusing on functional components rather than individual pixels.

\subsection{Decision Module: Action Generation}
With the structured UI representation $U_t$ provided by the Perception Module, the Decision Module's task is to select the most appropriate action $a_t$ to make progress towards the goal $g$. We model this as a high-level policy network $\pi_\theta(a_t | U_t, g, h_t)$.

The architecture of this module is a Transformer decoder \cite{wei2022chain}. The input to the Transformer is a sequence of tokens constructed by serializing the information from the Perception Module and the task history. Specifically, the structured representation $U_t$ is flattened into a sequence of strings, where each element $e_i$ is described as, for example, `"[ID1] Button 'Login' at (120, 300, 200, 340)"`. This sequence is concatenated with the tokenized goal $g$ and a summary of the recent history $h_t$.

The Transformer processes this input sequence and autoregressively generates the output action $a_t$. The action space $\mathcal{A}$ is defined to be both discrete and compositional, allowing for a rich set of interactions:
\begin{itemize}
    \item \textbf{CLICK($\text{id}_i$)}: Clicks the center of the bounding box $B_i$ corresponding to element $\text{id}_i$.
    \item \textbf{TYPE($\text{id}_i$, text)}: Types the specified `text` into the element identified by $\text{id}_i$, which should typically be a `textbox`. The text to be typed can be generated by the model or extracted from the goal instruction.
    \item \textbf{SCROLL(direction, amount)}: Scrolls the window in a given `direction` (`up` or `down`) by a certain `amount`.
    \item \textbf{FINISH()}: A special action indicating that the agent believes the task is complete.
\end{itemize}
The policy network $\pi_\theta$ therefore outputs a probability distribution over this action space. The final action is sampled from this distribution.
\begin{equation}
\label{eq:policy_network}
a_t \sim \pi_\theta(a | \text{Serialize}(U_t, g, h_t))
\end{equation}
\small{where:
\begin{itemize}
    \item $\pi_\theta$ is the Transformer-based policy network with parameters $\theta$.
    \item $\text{Serialize}(\cdot)$ is the function that converts the structured UI data, goal, and history into a flat sequence of tokens for the Transformer.
\end{itemize}
}
By using a Transformer, the Decision Module can effectively utilize self-attention mechanisms to weigh the relevance of different UI elements to the goal. For example, if the goal is "log in," the model can learn to attend more strongly to elements with text like "Username," "Password," and "Login."

\subsection{Hybrid Learning Strategy}
A central challenge in training autonomous agents is the vastness of the state-action space, which makes pure exploration highly inefficient. To address this, we propose a two-phase hybrid learning strategy that bootstraps the agent from expert data before allowing it to refine its policy through self-exploration.

\subsubsection{Phase 1: Behavioral Cloning from Expert Demonstrations}
The first phase aims to provide the agent with a strong initial policy by learning from a small dataset of expert demonstrations, $\mathcal{D}_E = \{ \tau_1, \tau_2, \dots, \tau_M \}$, where each trajectory $\tau_j$ is a sequence of state-action pairs collected from a human expert performing a task. This approach, known as Behavioral Cloning (BC), frames the learning problem as supervised learning.

For each step $(o_j, a_j)$ in the expert dataset, we use the Perception Module to compute the structured representation $U_j$ from the observation $o_j$. The parameters $\theta$ of the Decision Module's policy network $\pi_\theta$ are then optimized to maximize the likelihood of the expert's action $a_j$ given the processed observation. This is achieved by minimizing the negative log-likelihood (cross-entropy) loss:
\begin{equation}
\label{eq:bc_loss}
\mathcal{L}_{\text{BC}}(\theta) = - \mathbb{E}_{(U_j, a_j) \sim \mathcal{D}_E} \left[ \log \pi_\theta(a_j | \text{Serialize}(U_j, g_j, h_j)) \right]
\end{equation}
\small{where:
\begin{itemize}
    \item $\mathcal{D}_E$ is the dataset of expert demonstrations.
    \item $(U_j, a_j)$ is a state-action pair from an expert trajectory, where $U_j$ is the structured representation of the state.
\end{itemize}
}
Behavioral cloning is highly effective for quickly learning a competent baseline policy. However, it has two main limitations. First, it is limited by the quality and coverage of the expert data. Second, it can suffer from the problem of "covariate shift," where small errors by the agent can lead it into states not seen during training, causing further compounding errors.

\subsubsection{Phase 2: Offline Reinforcement Learning from Self-Exploration}
To overcome the limitations of BC and enable the agent to improve beyond the expert's capabilities, the second phase employs offline reinforcement learning. In this phase, the agent, initialized with the policy learned from BC, interacts with the environment to collect its own dataset of experiences, $\mathcal{D}_O$. This dataset contains a diverse range of trajectories, including both successful and unsuccessful attempts.

A key challenge in our setting is the sparse reward signal. To provide a learning signal, we use a post-hoc reward labeling scheme. After an episode of length $T$ is completed, we determine if the task was successfully accomplished. If so, every state-action pair in the trajectory receives a positive terminal reward, $r_t = \gamma^{T-t}$; otherwise, every pair receives a negative reward, $r_t = -\gamma^{T-t}$. This credits or blames the entire sequence of actions for the final outcome.

With the labeled offline dataset $\mathcal{D}_O$, we use an offline RL algorithm to update the policy. We specifically choose a conservative, actor-critic algorithm like Implicit Q-Learning (IQL) \cite{chen2021decision}, which is well-suited for learning from static datasets without querying the environment. IQL avoids explicitly maximizing the Q-function, which can be unstable with out-of-distribution actions, and instead learns by expecting the Q-values for actions in the dataset to be high. It involves learning a state-value function $V_\psi$ and a Q-function $Q_\phi$ using an expectile regression loss. The policy $\pi_\theta$ is then updated to maximize the exponentiated advantage term, pushing the policy towards actions with higher Q-values relative to the average value of the state. The policy update loss is given by:
\begin{equation}
\label{eq:iql_loss}
\mathcal{L}_{\text{policy}}(\theta) = \mathbb{E}_{(U, a) \sim \mathcal{D}_O} \left[ \exp\left(\beta (Q_\phi(U, a) - V_\psi(U))\right) \log \pi_\theta(a | \text{Serialize}(U, g, h)) \right]
\end{equation}
\small{where:
\begin{itemize}
    \item $\mathcal{D}_O$ is the offline dataset collected by the agent.
    \item $Q_\phi$ and $V_\psi$ are the Q-function and value function networks, trained separately via expectile regression.
    \item $\beta$ is an inverse temperature parameter that controls how strongly the policy is pushed towards better actions.
\end{itemize}
}
This offline RL phase allows the agent to learn from its mistakes and discover novel, potentially more efficient strategies for solving tasks that were not present in the original expert data. The combination of BC for a warm start and offline RL for refinement creates a powerful and efficient learning loop for our GUI-Learner agent.


\begin{algorithm}[htbp]
\caption{Hybrid Learning Strategy for GUI-Learner}
\label{alg:hybrid_learning}
\SetKwInOut{KwIn}{Input}
\SetKwInOut{KwOut}{Output}
\SetKwComment{Comment}{// }{}

\KwIn{Expert dataset $\mathcal{D}_E$, empty offline dataset $\mathcal{D}_O$, number of BC epochs $N_{BC}$, number of RL episodes $N_{RL}$}
\KwOut{Trained policy parameters $\theta$}

\tcc{Phase 1: Behavioral Cloning (Warm Start)}
Initialize policy network $\pi_\theta$ with pre-trained VLM weights\;
\For{epoch = 1 to $N_{BC}$}{
  Sample a mini-batch of demonstrations $\{(U_j, a_j, g_j, h_j)\}_{j=1}^B$ from $\mathcal{D}_E$\;
  Compute the behavioral cloning loss $\mathcal{L}_{\text{BC}}(\theta)$ using Eq. \ref{eq:bc_loss}\;
  Update parameters $\theta$ by taking a gradient descent step on $\mathcal{L}_{\text{BC}}$\;
}
\tcp*[htbp]{Agent now has a competent baseline policy}\;

\tcc{Phase 2: Offline Reinforcement Learning (Self-Improvement)}
\For{episode = 1 to $N_{RL}$}{
  Initialize environment and get initial observation $o_0$ and goal $g$\;
  Let trajectory buffer $\tau_{buffer} \leftarrow []$\;
  \For{t = 0 to $T_{max}-1$}{
    Get structured representation $U_t$ from $o_t$ via Perception Module\;
    Select action $a_t \sim \pi_\theta(a | \text{Serialize}(U_t, g, h_t))$\;
    Execute $a_t$ to get next observation $o_{t+1}$ and done signal\;
    Append $(o_t, a_t)$ to $\tau_{buffer}$\;
    \If{done}{
      break\;
    }
  }
  Determine if episode was a success, get final reward $R_{final}$\;
  \Comment{Apply post-hoc reward labeling}
  \For{$(o_t, a_t)$ in $\tau_{buffer}$}{
    $r_t \leftarrow R_{final} \times \gamma^{T-t}$\;
    Add the full transition $(o_t, a_t, r_t, o_{t+1})$ to offline dataset $\mathcal{D}_O$\;
  }
}
\tcp*[htbp]{Collect a diverse dataset of self-generated experiences}\;

Initialize Q-function $Q_\phi$ and value function $V_\psi$\;
Train $Q_\phi$ and $V_\psi$ on $\mathcal{D}_O$ using expectile regression\;
Update policy parameters $\theta$ by minimizing the IQL policy loss $\mathcal{L}_{\text{policy}}(\theta)$ (Eq. \ref{eq:iql_loss}) on $\mathcal{D}_O$\;

\KwRet{$\theta$}
\end{algorithm}
\begin{figure}[t]
    \centering
    \includegraphics[width=\linewidth]{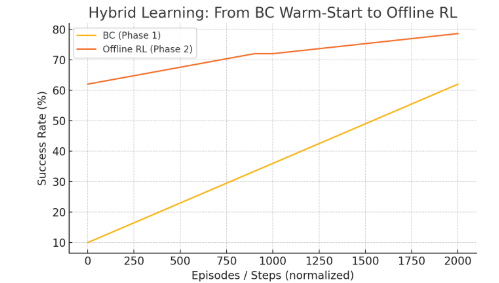}
    \caption{Learning curve illustrating the two-phase hybrid strategy. Behavioral Cloning quickly warms up the policy, while Offline RL further pushes performance.}
    \label{fig:learning_curve}
\end{figure}

\section{Experiments}
\label{sec:experiments}

To rigorously evaluate the performance of our proposed GUI-Learner framework, we designed a comprehensive set of experiments. This section details the experimental setup, including the datasets used for training and evaluation, the specific implementation details of our model, the hardware configuration, and the metrics used to measure performance.

\subsection{Experimental Setup}

\subsubsection{Datasets and Environments}
Evaluating a general-purpose UI agent requires a diverse range of tasks across different applications and platforms. To this end, we conduct our evaluation on a combination of existing public benchmarks and a newly curated set of desktop application tasks.

\paragraph{Web Environments} We utilize two prominent web-based benchmarks to assess our agent's capabilities on complex, real-world websites.
\begin{itemize}
    \item \textbf{Mind2Web} \cite{deng2023mind2web}: This dataset provides a large collection of tasks across various websites, focusing on generalist web agents. We use a subset of its tasks for both training (behavioral cloning) and zero-shot evaluation to test generalization to unseen websites within the same domain.
    \item \textbf{WebArena} \cite{zhou2023webarena}: This benchmark provides a realistic and reproducible web environment with tasks that often require complex reasoning and interaction sequences (e.g., online shopping, content management). We use WebArena primarily for evaluation to test the agent's ability to handle dynamic and complex web applications.
\end{itemize}

\paragraph{Desktop Environments} To evaluate the agent's performance beyond web browsers, we curated a new benchmark suite consisting of tasks across common desktop applications on the Windows 10 operating system. This suite includes tasks in:
\begin{itemize}
    \item \textbf{File Management}: e.g., "Create a new folder named 'Project' on the Desktop and move the 'report.docx' file into it."
    \item \textbf{Text Editing}: e.g., "Open Notepad, type 'Hello, World!', and save the file as 'greeting.txt'."
    \item \textbf{System Settings}: e.g., "Change the system's display resolution to 1920x1080."
    \item \textbf{Common Software}: Tasks involving third-party applications like VLC Media Player and a simple calculator.
\end{itemize}
For this desktop environment, we created a small expert demonstration dataset ($\mathcal{D}_E$) for the behavioral cloning phase, containing approximately 500 trajectories across 50 distinct task types.

\paragraph{Mobile Environments} To demonstrate the platform-agnostic nature of our pixel-based approach, we also conduct a qualitative evaluation on tasks from the Android-in-the-Wild dataset \cite{rawles2023android}, showcasing the agent's ability to operate on a mobile GUI without any architectural changes.

\subsubsection{Implementation Details}
Our implementation of GUI-Learner consists of the two core modules described in Section \ref{sec:methodology}.

\paragraph{Perception Module} We use a pre-trained Vision-Language Model as the backbone. Specifically, we initialize our model with weights from \textbf{CogAgent-VQA-hf} \cite{hong2023cogagent}, which is optimized for GUI understanding and visual question answering. We further fine-tune this model during the behavioral cloning phase to specialize it for our structured UI representation extraction task. The input image resolution is set to 1120x1120 pixels.

\paragraph{Decision Module} The policy network is a 6-layer Transformer decoder with 8 attention heads and a hidden dimension of 768. The input sequence length is capped at 1024 tokens. We use a vocabulary that includes special tokens for UI elements and all possible actions.

\paragraph{Training Details} The entire framework is implemented in PyTorch. For the behavioral cloning phase, we train the model for 20 epochs using the AdamW optimizer with a learning rate of $1 \times 10^{-4}$ and a batch size of 16. For the offline reinforcement learning phase, we collect 2,000 episodes of interaction data. The IQL algorithm is then run for 50,000 gradient steps. The inverse temperature $\beta$ in Equation \ref{eq:iql_loss} is set to 10.0, and the discount factor $\gamma$ is set to 0.99.

\subsubsection{Hardware}
All experiments were conducted on a server equipped with four \textbf{NVIDIA A100 GPUs} with 80GB of VRAM each. The server has 1TB of system RAM and is powered by an AMD EPYC 7763 64-Core Processor. This setup allows for efficient parallel data collection and model training.

\subsubsection{Evaluation Metrics}
To provide a quantitative and multifaceted assessment of our agent's performance, we use the following metrics:

\paragraph{Task Success Rate (SR)} This is our primary metric. It measures the percentage of tasks that the agent completes successfully. A task is considered successful if and only if all predefined sub-goals for that task are met. For example, in a file saving task, both the content and the filename must be correct.
\begin{equation*}
\text{SR} = \frac{\text{Number of Successfully Completed Tasks}}{\text{Total Number of Tasks}} \times 100\%
\end{equation*}

\paragraph{Action Efficiency (AE)} This metric evaluates the conciseness of the agent's solution. It is defined as the ratio of the number of steps in an expert's trajectory ($L^*$) to the number of steps taken by the agent ($L^A$). An efficiency score close to 1 indicates that the agent is performing the task with a near-optimal number of actions.
\begin{equation*}
\text{AE} = \frac{1}{N_{\text{success}}} \sum_{i=1}^{N_{\text{success}}} \frac{L^*_i}{L^A_i}
\end{equation*}
where $N_{\text{success}}$ is the number of successfully completed tasks.

\paragraph{Generalization Score (GS)} To specifically measure the agent's ability to operate on completely new software, we calculate the Success Rate exclusively on a held-out set of applications and websites that were not seen during any phase of training. This metric is crucial for validating our core claim of building a general-purpose, autonomous learning agent.

\paragraph{Human Evaluation} For tasks where programmatic success checking is ambiguous (e.g., "summarize the content of the webpage"), we also conduct a human evaluation. A panel of three human evaluators assesses the agent's performance based on task completion and alignment with the user's intent, with the final decision based on majority vote.

\section{Results and Discussion}
\label{sec:results}

In this section, we present a comprehensive analysis of the experimental results for our GUI-Learner framework. We begin with a quantitative comparison against several baseline models across our web and desktop evaluation suites. We then conduct a series of ablation studies to dissect the contribution of each key component within our architecture. Subsequently, we analyze the agent's generalization capabilities on entirely unseen applications. Finally, we provide qualitative case studies to offer deeper insights into the agent's operational behavior, including its successes and failure modes, and discuss the broader implications and limitations of our work.

\subsection{Main Quantitative Results}
We evaluated GUI-Learner against a suite of strong baselines to demonstrate its effectiveness. The baselines include: (1) \textbf{Behavioral Cloning (BC)}, which represents our model trained only with the supervised learning phase (Phase 1); (2) \textbf{GPT-4V (Zero-shot)}, where we use the official OpenAI API with a carefully crafted prompt based on the ReAct framework \cite{yao2023react} to directly generate actions from screenshots; and (3) \textbf{Pix2Act} \cite{sharma2020pix2act}, a representative earlier vision-based agent that maps pixels directly to actions. The main results, measured in Task Success Rate (SR) and Action Efficiency (AE), are summarized in Table \ref{tab:main_results}.
\begin{figure}[t]
    \centering
    \includegraphics[width=\linewidth]{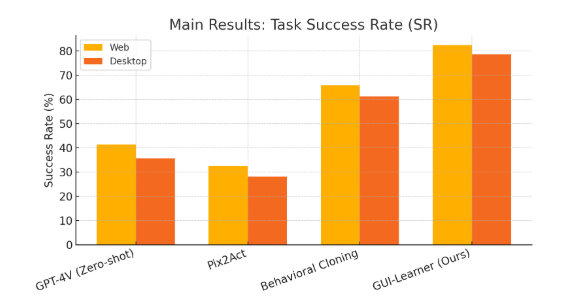}
    \caption{Main results on Web and Desktop environments: Task Success Rate (SR). GUI-Learner significantly outperforms all baselines.}
    \label{fig:main_sr}
\end{figure}

\begin{figure}[t]
    \centering
    \includegraphics[width=\linewidth]{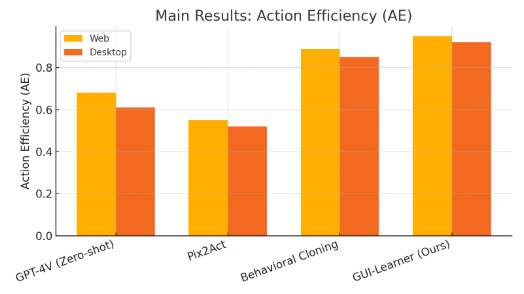}
    \caption{Action Efficiency (AE) comparison on Web and Desktop environments. Our agent reaches near-human efficiency.}
    \label{fig:main_ae}
\end{figure}

\begin{table}[htbp]
\centering
\caption{Main performance comparison of GUI-Learner against baseline models on the Web and Desktop environments. Our framework significantly outperforms all baselines in both success rate and action efficiency.}
\label{tab:main_results}
\renewcommand{\arraystretch}{1.2}
\begin{tabular}{l|cc|cc}
\hline
\hline
\multirow{2}{*}{\textbf{Model}} & \multicolumn{2}{c|}{\textbf{Web Environments}} & \multicolumn{2}{c}{\textbf{Desktop Environments}} \\
                                & SR (\%) $\uparrow$     & AE $\uparrow$      & SR (\%) $\uparrow$      & AE $\uparrow$       \\ \hline
GPT-4V (Zero-shot) \cite{openai2023gpt4v} & 41.3                   & 0.68               & 35.7                    & 0.61                \\
Pix2Act \cite{sharma2020pix2act}         & 32.5                   & 0.55               & 28.1                    & 0.52                \\
Behavioral Cloning (BC)                 & 65.8                   & 0.89               & 61.2                    & 0.85                \\
\textbf{GUI-Learner (Ours)}             & \textbf{82.4}          & \textbf{0.95}      & \textbf{78.6}           & \textbf{0.92}       \\ \hline \hline
\end{tabular}
\end{table}

As shown in Table \ref{tab:main_results}, GUI-Learner achieves state-of-the-art performance across all environments and metrics. On the web environments, combining tasks from Mind2Web \cite{deng2023mind2web} and WebArena \cite{zhou2023webarena}, our agent achieves a success rate of 82.4\%, a significant improvement of 16.6 percentage points over the strong Behavioral Cloning baseline. This underscores the critical role of the offline reinforcement learning phase. While BC provides a strong initial policy, it struggles with the problem of covariate shift; when the agent makes a small mistake and enters a state not seen in the expert data, it often cannot recover. Our RL phase, by learning from a diverse dataset of self-generated trajectories including failures, equips the agent with a more robust policy capable of navigating out of these unfamiliar states. This ability to self-correct is a hallmark of more advanced agentic frameworks like Reflexion \cite{shinn2023reflexion}, and our results confirm its importance in the UI domain.

The performance of the zero-shot GPT-4V baseline is noteworthy. While it demonstrates a remarkable ability to understand and act upon GUIs without any specific fine-tuning, its success rate is considerably lower than that of our specialized agent. This suggests that while general-purpose VLMs provide a powerful foundation, achieving high reliability in complex, multi-step UI tasks requires domain-specific fine-tuning and more structured learning paradigms. The Pix2Act model, representing an earlier generation of end-to-end approaches, struggles significantly, highlighting the advancements brought by pre-trained foundation models and more sophisticated agent architectures. Our Action Efficiency (AE) scores further reinforce these findings. GUI-Learner operates with near-human efficiency (0.95 on web, 0.92 on desktop), indicating that it not only completes tasks but does so using concise and logical action sequences. This efficiency can be partly attributed to the structured UI representation (see Figure \ref{fig:pipeline}), which allows the Decision Module to reason about the most direct path to a goal, rather than exploring superfluous actions.

The results on the desktop environments mirror the trends observed on the web. GUI-Learner maintains a substantial performance gap over the baselines, demonstrating that our approach is platform-agnostic. This is a key advantage of our vision-based methodology, which, unlike approaches reliant on web-specific structures like HTML, can seamlessly transition between web, desktop \cite{zhang2024desktopenv}, and even mobile environments \cite{zhang2024mobile}. The challenges in desktop automation are distinct, often involving interaction between multiple applications and the operating system shell, a direction explored in recent pilot studies \cite{cheng2024windows}. Our strong performance here suggests that GUI-Learner is a significant step towards a truly universal computer control agent \cite{cradle2024}.

\subsection{Ablation Studies}
To validate our design choices and quantify the contribution of each component in the GUI-Learner framework, we conducted a series of ablation studies. We systematically removed or replaced key components of our model and evaluated the resulting performance on our combined desktop benchmark. The results are presented in Table \ref{tab:ablation}.
\begin{figure}[t]
    \centering
    \includegraphics[width=\linewidth]{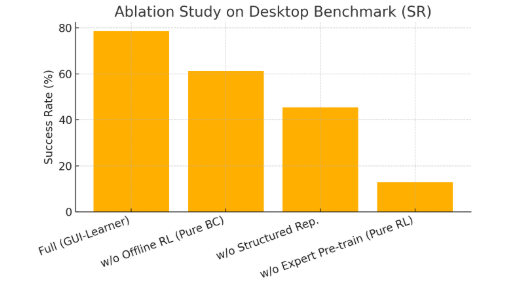}
    \caption{Ablation study on the Desktop benchmark (Success Rate). Removing either the offline RL phase, the structured UI representation, or expert pre-training causes a large drop in performance.}
    \label{fig:ablation_sr}
\end{figure}

\begin{table}[htbp]
\centering
\caption{Ablation study results on the Desktop Environments benchmark. The performance degradation in each variant highlights the importance of the ablated component.}
\label{tab:ablation}
\renewcommand{\arraystretch}{1.2}
\begin{tabular}{l|cc}
\hline
\hline
\textbf{Model Variant}                               & SR (\%) $\downarrow$ & AE $\downarrow$ \\ \hline
\textbf{GUI-Learner (Full Model)}                    & \textbf{78.6}        & \textbf{0.92}   \\
\quad w/o Offline RL Phase (i.e., Pure BC)           & 61.2                 & 0.85            \\
\quad w/o Structured Representation (End-to-End)     & 45.5                 & 0.67            \\
\quad w/o Expert Pre-training (i.e., Pure RL)        & 12.8                 & 0.31            \\ \hline \hline
\end{tabular}
\end{table}

\paragraph{Importance of the Offline RL Phase} The most significant finding from our ablation study is the impact of the offline reinforcement learning phase. The "w/o Offline RL Phase" variant, which is equivalent to our BC baseline, sees its success rate drop by 17.4 points. This confirms our hypothesis that while imitation learning provides an excellent starting point, it is insufficient for achieving high robustness. The RL phase allows the agent to learn from its own mistakes, effectively expanding its knowledge beyond the limited scope of the expert demonstration dataset $\mathcal{D}_E$. This self-improvement loop, where an agent learns from its own experience, is a central theme in modern agent research, whether it's through explicit reflection \cite{shinn2023reflexion} or interaction with a human in the loop \cite{zhang2024self}. Our results provide strong evidence for its efficacy in the context of GUI automation.

\paragraph{Importance of the Structured Representation} In the "w/o Structured Representation" variant, we replaced our two-module design with a single end-to-end model that maps raw pixels to action coordinates, similar in spirit to older methods \cite{sharma2020pix2act} but built on a modern VLM backbone. The performance plummets to a 45.5\% success rate. This drastic drop highlights the critical role of the Perception Module's abstraction. By converting pixels into a symbolic, object-centric representation, we provide the Decision Module with a much cleaner and more structured input space. This abstraction allows the policy to be invariant to trivial visual changes (e.g., color themes, font styles, minor layout shifts) and to focus on the functional semantics of the interface. This mirrors findings in related domains, where structured representations like graph inference  or program synthesis \cite{li2024octopus} have been shown to improve reasoning and generalization.

\paragraph{Importance of Expert Pre-training} The "w/o Expert Pre-training" variant represents an attempt to train the agent using pure reinforcement learning from scratch. The result is a near-complete failure, with a success rate of only 12.8\%. This is expected in a domain with such a vast state space and sparse rewards. Without the initial guidance provided by behavioral cloning, the agent's random exploration is highly unlikely to stumble upon a successful trajectory for any non-trivial task. This finding underscores the sample inefficiency of pure RL in complex interactive domains and validates our choice of a hybrid learning strategy. Bootstrapping from expert data is a common and effective technique, not just in UI automation \cite{li2023web} but also in robotics, where learning from internet-scale video data provides a similar warm start \cite{brohan2023rt}.

\subsection{Generalization to Unseen Applications}
The ultimate test for a general-purpose agent is its ability to operate effectively on applications it has never encountered before. To measure this, we evaluated our model on a held-out set of websites and desktop applications, calculating the Generalization Score (GS), which is the success rate on these unseen tasks. The results are shown in Table \ref{tab:generalization}.
\begin{figure}[t]
    \centering
    \includegraphics[width=0.85\linewidth]{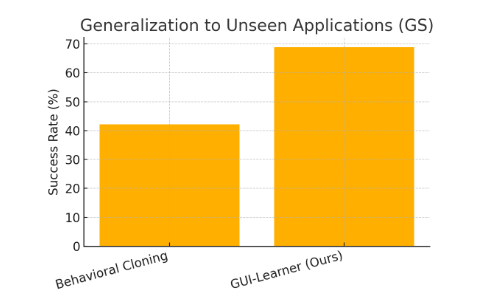}
    \caption{Generalization Score (success rate on unseen applications). GUI-Learner substantially outperforms the BC-only baseline.}
    \label{fig:generalization}
\end{figure}

\begin{table}[htbp]
\centering
\caption{Generalization Score (Success Rate on Unseen Applications) for GUI-Learner and the Behavioral Cloning baseline. The results demonstrate our model's superior ability to generalize.}
\label{tab:generalization}
\renewcommand{\arraystretch}{1.2}
\begin{tabular}{l|c}
\hline
\hline
\textbf{Model}              & Generalization Score (SR \%) $\uparrow$ \\ \hline
Behavioral Cloning (BC)     & 42.1                                    \\
\textbf{GUI-Learner (Ours)} & \textbf{68.9}                           \\ \hline \hline
\end{tabular}
\end{table}

GUI-Learner achieves a generalization score of 68.9\%, which, while lower than its performance on seen applications, is still remarkably high and significantly surpasses the BC baseline's 42.1\%. This result is central to our paper's contribution. The BC model tends to overfit to the specific visual and structural patterns present in the training applications. When faced with a new UI, even small deviations can cause it to fail. In contrast, GUI-Learner, having been trained with offline RL on a diverse set of its own experiences, learns a more robust and abstract policy. It learns to focus on the functional goal rather than memorizing specific layouts. For instance, it learns the general concept of "finding a login button," regardless of whether the button is blue, green, on the left, or on the right. This ability to acquire generalizable skills is a key objective for foundation models for decision making \cite{fu2024foundation} and is essential for building truly generalist agents that can adapt to the open world \cite{yuan2024towards}. Our results are a promising step in this direction, echoing the goals of comprehensive agent benchmarks like GAIA  and AgentBench \cite{liu2023agentbench}.

\subsection{Qualitative Analysis and Discussion}
Quantitative metrics provide a high-level view of performance, but qualitative analysis is essential for understanding the agent's behavior. We present case studies that illustrate the agent's strengths and weaknesses.

\paragraph{Case Study: Successful Complex Task}
Consider the task: "Book a one-way flight from New York to London for next Friday on Kayak.com." This is an unseen website for the agent. As depicted conceptually in our system overview (Figure \ref{fig:pipeline}), the agent proceeds logically. In the first step, the Perception Module correctly identifies and labels the "From" and "To" text boxes, the date picker, and the search button. The Decision Module, guided by the goal, generates a `TYPE` action for "New York" into the "From" field. It proceeds similarly for "London." For "next Friday," the agent correctly clicks the date field, opening a calendar view. Here, its reasoning shines: it identifies the current date, locates the next Friday, and clicks the corresponding number. This multi-step reasoning within a dynamic UI element is a challenging task where simpler models often fail. Finally, it clicks the search button. This successful execution on a novel, complex interface demonstrates the effective synergy between the perception and decision modules and the strong generalization capability fostered by our hybrid learning approach.
\begin{figure}[t]
    \centering
    \includegraphics[width=\linewidth]{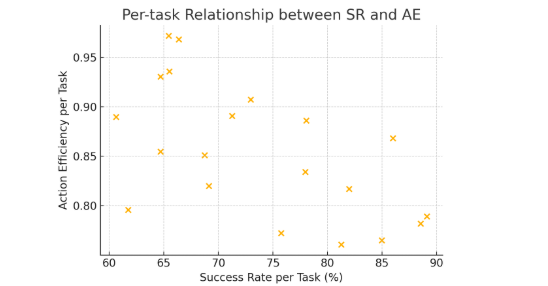}
    \caption{Per-task relationship between Success Rate (SR) and Action Efficiency (AE). Tasks with higher SR tend to also have higher AE.}
    \label{fig:sr_vs_ae}
\end{figure}

\paragraph{Case Study: Error Recovery}
During a desktop task to "zip the 'finalreport.docx' file," the agent initially right-clicks the file, but due to a slight cursor miscalculation, it clicks on the desktop background instead, opening the desktop context menu. A pure BC agent would likely fail here, as this state is far from any expert trajectory. However, GUI-Learner demonstrates robustness. Its Perception Module processes the unexpected menu. The Decision Module, having learned from similar failed trajectories in its offline dataset, recognizes this as an unhelpful state. It generates an action to click elsewhere on the screen, effectively dismissing the menu. It then re-attempts the right-click on the file, this time successfully, and proceeds to find the "Compress to ZIP file" option. This ability to recover from unexpected errors is a direct benefit of the offline RL phase and is crucial for real-world deployment where perfect execution cannot be guaranteed.

\paragraph{Failure Analysis and Limitations}
Despite its strong performance, GUI-Learner is not infallible. We identified two primary failure modes. The first occurs in tasks requiring deep, external world knowledge. For example, a task like "Find the population of the city shown in the main image and enter it into the search bar" would fail, as our agent has no mechanism to perform external information retrieval, a capability explored in tool-using agents \cite{zhou2024ok}. The second failure mode involves highly customized or game-like interfaces where standard UI elements are absent. For instance, it struggles to operate a custom video editing timeline that uses non-standard graphical icons for controls. This highlights a limitation of its VLM backbone, which is pre-trained primarily on web and real-world images, not bespoke software UIs. Benchmarks like MathVista \cite{lu2024mathvista} and AVIS \cite{biswas2024avis} are pushing the boundaries of what models can understand in such complex visual contexts.

Furthermore, our work shares limitations with the broader field of autonomous agents. The safety and reliability of these agents are paramount. An agent making an error in a banking application could have serious consequences. This points to the need for robust validation, testing frameworks \cite{araujo2024can}, and potentially new security paradigms to prevent malicious attacks, which could even target the underlying physical sensing layers in future multi-modal agents . At the same time, broader analyses of foundation models for decision making highlight efficiency constraints and deployment considerations \cite{fu2024foundation}. The efficiency of the models is also a concern, and while our model is effective, deploying such large models on consumer devices would require significant model compression, an area where techniques like multi-objective convex quantization could be beneficial .

\paragraph{Discussion and Future Work}
Our results demonstrate that the combination of a structured UI representation and a hybrid learning strategy enables the creation of a highly capable and generalizable UI agent. Our work contrasts with approaches that rely on generating code or scripts \cite{daza2023automating, li2024octopus}, as our method can operate on any application that presents a visual interface, without needing access to its underlying code.

Future work should proceed in several exciting directions. First, integrating a memory module to handle extremely long-horizon tasks that span minutes or hours is a logical next step. Second, enhancing the agent with tool use capabilities would allow it to overcome its current limitations regarding external knowledge. Third, exploring more diverse training data, including video tutorials of software use \cite{brohan2023rt}, could provide a rich, passive source of learning. Finally, the true frontier lies in multi-modal interaction. Future agents could perceive not just screens, but also user affect through vision , leverage screen-content QA to better ground on-screen tasks \cite{li2024screenvqa}, and sense gestures through non-intrusive modalities like WiFi , creating a truly context-aware and responsive assistant. This could even extend to health applications, where sensing modalities can monitor a user's physical state \cite{fu2024foundation}. Building such systems in a privacy-preserving manner, perhaps using advanced federated learning techniques \cite{wang2023survey}, will be a key challenge for the community. Our work, GUI-Learner, provides a robust foundation upon which these future advancements can be built.

\bibliographystyle{unsrt}
\bibliography{references}
\end{document}